\documentclass{article}

\PassOptionsToPackage{numbers}{natbib}

 \usepackage[preprint]{neurips_2026}

\usepackage[utf8]{inputenc} % allow utf-8 input
\usepackage[T1]{fontenc}    % use 8-bit T1 fonts
\usepackage{hyperref}       % hyperlinks
\usepackage{url}            % simple URL typesetting
\usepackage{booktabs}       % professional-quality tables
\usepackage{amsfonts}       % blackboard math symbols
\usepackage{nicefrac}       % compact symbols for 1/2, etc.
\usepackage{microtype}      % microtypography
\usepackage[dvipsnames]{xcolor}         % colors
\usepackage{dashrule}       % for \hdashrule
\usepackage{enumitem}
\usepackage{wrapfig}
\title{\dataset: A Benchmark for Spatiotemporal Intelligence in VLMs for Autonomous Driving}

\author{%
Hao Vo$^{1}$,
Khoa Vo$^{1}$,
Phu Loc Nguyen$^{1}$,
Sieu Tran$^{1}$,
Duc Minh Nguyen$^{1}$,
Ngo Xuan Cuong$^{1}$, \\
\bf Gladys Gawugah$^{1}$,
 Sreevenkata Anjani Tishita Godavarthi$^{1}$,
Chase Rainwater$^{1}$,
Nghi D. Q. Bui$^{2}$, \\
\bf Anh Nguyen$^{3}$,
Duy Minh Ho Nguyen$^{4}$,
Ngan Le$^{1}$\\
$^{1}$University of Arkansas, USA \quad
$^{2}$Center of AI Research, VinUniversity, Viet Nam \quad \\
$^{3}$University of Liverpool, UK \quad
$^{4}$Max Planck Research School for Intelligent Systems \quad \\
\small{https://uark-aicv.github.io/DriveSpatial/} \\
\vspace{-6mm}
}

\usepackage{xspace}
\usepackage{graphicx}
\usepackage{multirow}
\usepackage{array}
\usepackage{booktabs}        % cleaner rules (not mandatory but recommended)
\usepackage{makecell}        % for \rotatebox + cell formatting
\usepackage{tabularx}
\usepackage{pifont}

\usepackage{graphicx}        % \scalebox, \rotatebox
\usepackage{caption}         % \captionsetup{type=table}
\usepackage{subcaption}      % safe to include with caption
\usepackage{pdfpages}

\usepackage[table]{xcolor}   % \rowcolor, \cellcolor, custom colors

\usepackage{fontsize}        % optional, for \fontsize (often already supported)

\usepackage{amsmath}
\usepackage{amssymb}
\usepackage{xcolor}
\usepackage{booktabs}
\usepackage{multirow}
\usepackage{graphicx}   % for \rotatebox
\usepackage{makecell}   % optional, for centering text in cells
\usepackage{float}
\usepackage[most]{tcolorbox}
\usepackage{soul}

\tcbset{
  adbox/.style={
    breakable,
    enhanced,
    colback=gray!10,
    colframe=gray!50!black,
    arc=2mm,
    boxrule=0.8pt,
    left=6pt,
    right=6pt,
    top=6pt,
    title={\textcolor{white}{\bfseries\Large #1}},
    bottom=6pt,
    overlay unbroken={},    % Add these 4 lines
    overlay first={},       % to remove
    overlay middle={},      % the dots
    overlay last={}         % completely
  }
}
\usepackage[toc,page,header]{appendix}
\usepackage{minitoc}
\usepackage{algorithm}
\usepackage{algpseudocode}

\definecolor{navyblue}{RGB}{230,235,245}

\definecolor{oai-gray-300}{RGB}{210,210,210}
\definecolor{oai-gray-600}{RGB}{160,160,160}

\definecolor{oai-green-200}{RGB}{198,239,206}
\definecolor{oai-green-400}{RGB}{123,201,111}
\definecolor{oai-green-600}{RGB}{0,176,80}
\definecolor{lightblue}{rgb}{0.9,0.95,1.0} 
\definecolor{lightgray}{RGB}{240, 243, 246}
\usepackage{xcolor}
\usepackage[most]{tcolorbox}
\definecolor{lightgraybg}{gray}{0.95}
\definecolor{darkgreen}{RGB}{0,128,0} 
\usepackage{forest}
\usepackage{dirtree}

\definecolor{appendixlink}{RGB}{142, 178, 230}

\newcommand{\dataset}{\textsc{DriveSpatial}\xspace}

\definecolor{abilityCog}{RGB}{231,76,60}
\definecolor{abilityMV}{RGB}{52,152,219}
\definecolor{abilityTmp}{RGB}{243,156,18}
\definecolor{abilityGen}{RGB}{39,174,96}

\newcommand{\ulCognitiveScene}{\setulcolor{abilityCog}\ul{\textit{Cognitive Scene Construction}}}

\newcommand{\ulMultiViewRel}{\setulcolor{abilityMV}\ul{\textit{Multi-view Relational Understanding}}}
\newcommand{\ulTemporalReason}{\setulcolor{abilityTmp}\ul{\textit{Temporal Reasoning}}}

\newcommand{\ulGeneralization}{\setulcolor{abilityGen}\ul{\textit{Generalization}}}

\begin{document}
% Required for minitoc: collects headings so \parttoc is not empty (run pdflatex twice).
\doparttoc
\faketableofcontents

\maketitle

\begin{figure}[ht]
  \centering
  \includegraphics[page=1, width=\linewidth]{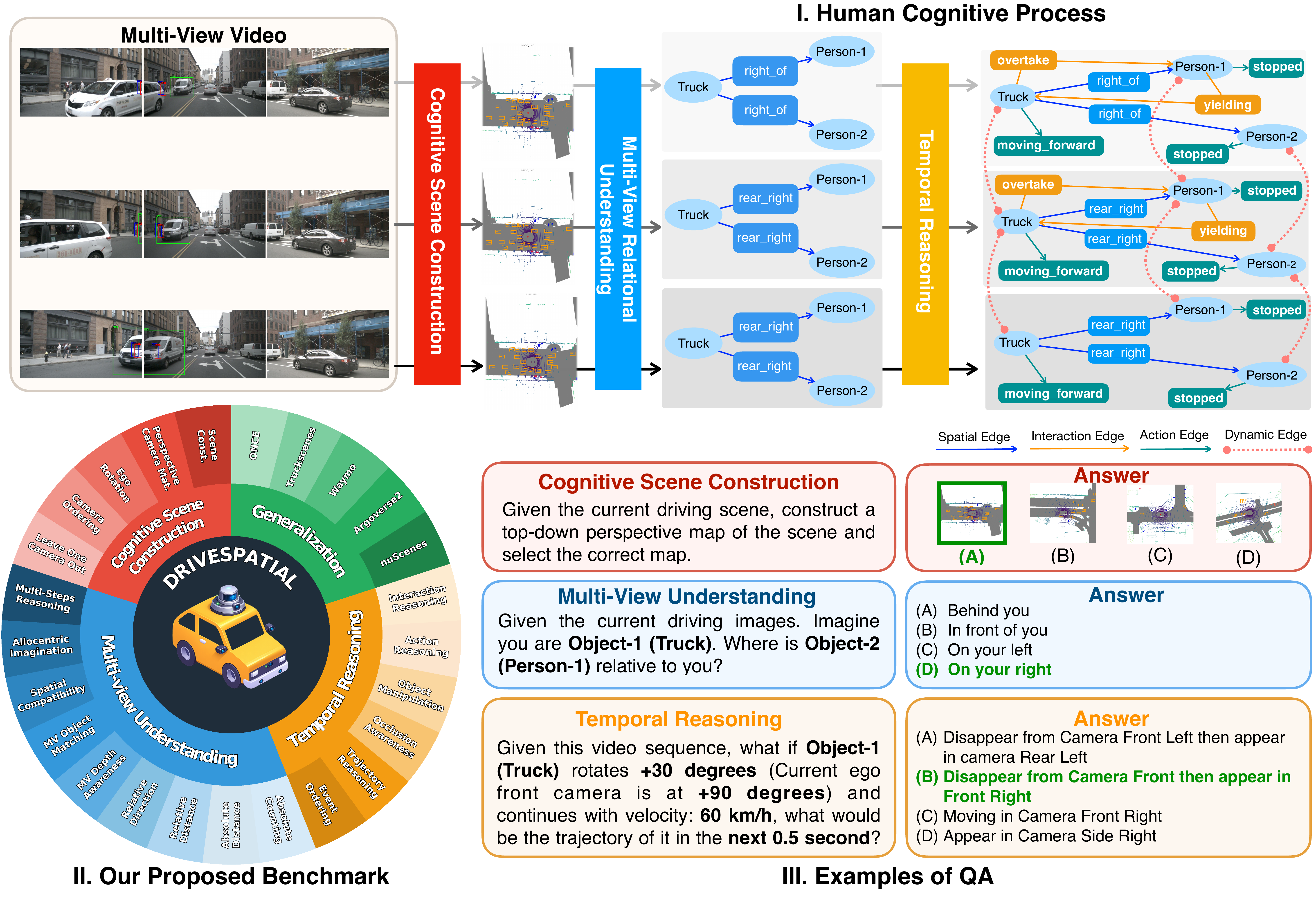}
    % \caption{\textbf{Overview of \dataset{}.} \textit{(I, Top)} Spatiotemporal intelligence in driving mirrors the human cognitive process: multi-view video observations are first  are first used to mentally construct an internal representation (\textit{Scene Construction}), from which spatial relationships between objects are inferred (\textit{Multi-view Relational Understanding}), and finally connected across time to support causal inference and anticipation (\textit{Temporal Reasoning}). \textit{(II, Bottom left)} \dataset{} covers 18 tasks organized under four core abilities, sourced from five AV datasets. \textit{(III, Bottom right)} Representative questions from three ability categories; correct answers are highlighted in green. \textcolor{red}{ Can not see the bounding box, the text look all black, add the place holder to the questions}
  % }
  \caption{
  We present \textbf{\dataset{}}: A spatiotemporal intelligence evaluation benchmark for Autonomous Driving that mirrors human navigation cognition. \textit{(I, Top)} In driving scenarios, humans gather observations from multiple viewpoints to mentally construct an internal representation (\textit{\protect\ulCognitiveScene}), infer spatial relationships between objects (\textit{\protect\ulMultiViewRel}), and connect these percepts across time to support causal inference and anticipation (\textit{\protect\ulTemporalReason}). \textit{(II, Bottom left)} \textbf{\dataset{}} encompasses 20 tasks organized under four core cognitive abilities, derived from five distinct AD datasets. \textit{(III, Bottom right)} Representative questions spanning three ability categories are shown, with correct answers highlighted in \textcolor{OliveGreen}{green}.
  }
\end{figure}

\begin{abstract}

Spatiotemporal intelligence in autonomous driving (AD) requires an agent to integrate multi-view observations into a coherent scene representation, maintain object continuity across viewpoints and time, and reason about spatial relations, interactions, and future dynamics. However, existing AD vision-language benchmarks largely focus on single-view, static, ego-centric, or single-source question answering, leaving it unclear whether current Vision-Language Models (VLMs) can truly construct and reason over dynamic driving scenes. We introduce \textbf{\dataset} a benchmark of 15.6K human-verified QA pairs across 20 tasks from five large-scale AD datasets. DRIVESPATIAL evaluates four abilities: \textit{\protect\ulCognitiveScene}, \textit{\protect\ulMultiViewRel}, \textit{\protect\ulTemporalReason}, and \textit{\protect\ulGeneralization}. Unlike prior benchmarks, DRIVESPATIAL is generated from a dynamic multi-relational scene graph that encodes object states, spatial relations, interactions, camera visibility, and temporal correspondences, enabling QA pairs that enforce genuine cross-view and spatiotemporal reasoning. Evaluating 15 representative VLMs reveals a substantial human–model gap: the strongest model trails humans by 28.4 points, with \textit{\protect\ulCognitiveScene} emerging as the key bottleneck. Further diagnostics show that language-only prompting is insufficient, while explicit BEV grounding consistently improves performance. These results suggest that current VLMs lack the scene-construction ability needed for reliable spatiotemporal driving intelligence. DRIVESPATIAL and its construction pipeline will be released to support future research.

\end{abstract}

\section{Introduction}
%\nghi{I rewrote the introduction to sharpen the novelty claim: DriveSpatial is now positioned as a benchmark for coherent multi-view scene construction, cross-view reasoning, and temporal reasoning, rather than general driving QA.}
\textit{Spatiotemporal Intelligence} refers to the ability to mentally construct, understand, and reason about an internal representation from the surrounding environment. It is a fundamental component of human cognition for navigation, particularly in driving scenarios~\cite{tolman1946studies, tolman1948cognitive, nadel2008hippocampus, burnett2005effect}. By integrating information from multiple viewpoints, humans form a coherent spatial representation that supports relational reasoning, occlusion understanding, and risk anticipation. As vision–language models (VLMs) become increasingly embedded in Autonomous Driving (AD)~\cite{wang2025omnidrive,li2024ego,ding2024holistic,xie2025vlms,xu2024vlm,huang2025robotron,cao2025fastdrivevla,xu2024drivegpt4}, evaluating models with comparable spatial reasoning capabilities becomes essential for robust perception, interaction, and decision-making.

\noindent
Recent VLM-based AD systems and benchmarks have made substantial progress in language-grounded perception, prediction, and planning~\cite{arai2025covla, li2024ego, hwang2024emma, wang2025omnidrive, xu2024drivegpt4, sima2024drivelm, tian2024drivevlm, zeng2025futuresightdrive}. However, current evaluations still leave a critical question unresolved: \emph{do these models truly understand the spatial structure and temporal dynamics of a driving scene, or do they mostly rely on local recognition cues and dataset-specific regularities?} Most existing benchmarks rely heavily on a single dataset such as nuScenes~\cite{tian2025nuscenes}, focus on static scenes~\cite{cao2024maplm, tian2025nuscenes, gholami2025spatial, guo2025surds, ding2024holistic, chen2025automated, marcu2024lingoqa, wu2025towards}, restrict reasoning to a single camera view~\cite{marcu2024lingoqa, cao2024maplm,guo2025surds, tian2025nuscenes, chen2025automated, sima2024drivelm, wu2025towards}, or offer limited question diversity~\cite{gholami2025spatial, fruhwirth2025stsbench}. As a result, strong benchmark performance may overestimate a model’s true scene-level reasoning ability.

\noindent 
This gap is clearer when comparing what current benchmarks actually test. nuScenes-QA and related driving QA datasets remain largely ego-centric and do not explicitly enforce cross-view integration~\cite{qian2024nuscenes, tian2025nuscenes}. DriveLM and OmniDrive broaden driving-oriented QA, but their evaluations are still centered on perception, prediction, and planning rather than testing whether a model can reconstruct a coherent scene from synchronized camera streams~\cite{sima2024drivelm, wang2025omnidrive}. Ego3D-Bench and STSBench move closer to spatial and spatiotemporal evaluation, yet they still do not jointly probe scene construction, cross-view identity and relation reasoning, and temporal interaction reasoning under a single benchmark~\cite{gholami2025spatial, fruhwirth2025stsbench}. In addition, many existing questions can still be solved as conventional 2D visual question answering~\cite{guo2025surds, ding2024holistic, wang2025omnidrive, tian2025nuscenes}, without requiring a coherent multi-view scene representation.

\noindent
We argue that this mismatch comes from a gap between current benchmark design and real driving cognition. Safe driving requires building a coherent scene representation from multiple viewpoints and time steps, using that representation to interpret relations and interactions, and maintaining those abilities across new sources and conditions. These capabilities are central to spatiotemporal intelligence, but are still only partially measured by existing evaluation frameworks.

\begin{table}[t]
\centering
\caption{\textbf{Benchmark Comparison. }Existing datasets mainly evaluate whether a model can answer driving-related visual questions, whereas \dataset{} evaluates whether a model can construct, maintain, and use a coherent multi-view spatiotemporal scene representation. $^\dagger$ denotes the exclusion of questions related to ``Detect all objects in the current scene'' and any questions involving a single view.
Multi-view Constraint: whether question generation explicitly enforces cross-view reasoning.}
\label{tab:benchmarks_comparison}
\renewcommand{\arraystretch}{1.0}
\setlength{\tabcolsep}{6pt}
\resizebox{\linewidth}{!}{\large
\begin{tabular}{l|c|cccc|cc|cccc}
\toprule
\multirow{3}{*}{\textbf{Name}} &
\multirow{3}{*}{{\textbf{Data} \textbf{Source}}} &
\multicolumn{4}{c|}{\textbf{Data Property}} &
\multicolumn{2}{c|}{\textbf{Question Property}} &
\multicolumn{4}{c}{\textbf{Spatiotemporal Intelligence }} \\
\cmidrule(lr){3-6}\cmidrule(lr){7-8}\cmidrule(lr){9-12}
& &
\textbf{\# Views} &
\makecell[c]{\textbf{Ego}\\\textbf{Type}} &
\makecell[c]{\textbf{\# QA}\\\textbf{Pairs}} &
\textbf{\# Tasks} &
\makecell[c]{\textbf{Multi-view}\\\textbf{Constraint}} &
\makecell[c]{\textbf{Question}\\\textbf{Type}} &
\cellcolor{red!10} \textbf{Const.} &
\cellcolor{blue!10} \textbf{Unders.$\dagger$} &
\cellcolor{yellow!10} \textbf{Reas.} &
\cellcolor{green!10} \textbf{Gen.} \\
\midrule
\rowcolor{lightgray}
\multicolumn{12}{l}{\emph{\textcolor{black}{Auto}}}\\
NuScenes-SpatialQA & nuScenes & 6 & Car & 3.5M & 14
& \textcolor{red}{\ding{55}} & Ego & \textcolor{red}{\ding{55}} & \textcolor{red}{\ding{55}} & \textcolor{red}{\ding{55}} & Low \\
nuScenes-QA & nuScenes & 6 & Car & 83.3k & 5 &
\textcolor{red}{\ding{55}} & Ego & \textcolor{red}{\ding{55}} & \textcolor{red}{\ding{55}} & \textcolor{red}{\ding{55}} & Low \\
NuInstruct & nuScenes & 6 & Car & 91k & 17 &
\textcolor{red}{\ding{55}} & Ego & \textcolor{red}{\ding{55}} & \textcolor{red}{\ding{55}} & \textcolor{darkgreen}{\ding{51}} & Low \\
DriveLMM-o1 & nuScenes & 6 & Car & 18k & 10 &
\textcolor{red}{\ding{55}} & Ego & \textcolor{red}{\ding{55}} & \textcolor{red}{\ding{55}} & \textcolor{darkgreen}{\ding{51}} & Low \\
\midrule
\rowcolor{lightgray}
\multicolumn{12}{l}{\emph{\textcolor{black}{Human-In-The-Loop}}}\\
OmniDrive & nuScenes & 6 & Car & -- & 5 &
\textcolor{red}{\ding{55}} & Ego & \textcolor{red}{\ding{55}} & \textcolor{red}{\ding{55}} & \textcolor{darkgreen}{\ding{51}} & Low \\
DriveLM & nuScenes & 6 & Car & 15.4k & 18 &
\textcolor{red}{\ding{55}} & Ego & \textcolor{red}{\ding{55}} & \textcolor{red}{\ding{55}} & \textcolor{red}{\ding{55}} & Low \\
DriveBench & nuScenes & 6 & Car & 20.5k & 18 &
\textcolor{red}{\ding{55}} & Ego & \textcolor{red}{\ding{55}} & \textcolor{red}{\ding{55}} & \textcolor{red}{\ding{55}} & Low \\
STSBench & nuScenes & 6 & Car & 971 & 4 &
\textcolor{red}{\ding{55}} & Ego+Allo & \textcolor{red}{\ding{55}} & \textcolor{red}{\ding{55}} & \textcolor{darkgreen}{\ding{51}} & Low \\
Ego3D-Bench  & \makecell{nuScenes, \\Waymo, AV1} & 5,6,7 & Car & 8.6k & 8 &
-- & Ego & \textcolor{red}{\ding{55}} & \textcolor{darkgreen}{\ding{51}} & \textcolor{red}{\ding{55}} & Medium \\
\rowcolor{lightblue}
\textbf{\shortstack{\dataset}} &
\makecell{nuScenes, \\ Waymo, AV2,\\TruckScenes, ONCE} &
4,5,6,7 &
Car/Truck &
15.6k & 20 &
Graph  & Ego+Allo & \textcolor{darkgreen}{\ding{51}} & \textcolor{darkgreen}{\ding{51}} & \textcolor{darkgreen}{\ding{51}} & High \\
\bottomrule
\end{tabular}
}
\vspace{-6mm}
\end{table}

\noindent
To fill these gaps, we introduce \textbf{\dataset}, a benchmark for evaluating spatiotemporal intelligence in AD. {\dataset} is derived from five large-scale AD benchmarks~\cite{caesar2020nuscenes,wilson2023argoverse,fent2024man,sun2020scalability,mao2021one} and contains 15.6k human-verified question-answer pairs spanning 20 tasks. While prior benchmarks rely on multi-view images alone and leave cross-view reasoning implicit, \dataset constructs its questions from a dynamic multi-relational scene graph that encodes object states, spatial relations, interactions, actions, camera visibility, and temporal correspondences across frames. The benchmark is organized around three core reasoning abilities and one robustness axis: \textbf{(1) \ulCognitiveScene} (\colorbox{red!10}{Const.}), the ability to build an internal representation of the surrounding environment from multi-view observations; \textbf{(2) \ulMultiViewRel} (\colorbox{blue!10}{Unders.}), the ability to interpret spatial relationships among objects across multiple viewpoints; \textbf{(3) \ulTemporalReason} (\colorbox{yellow!10}{Reas.}), the ability to connect observations over time to infer causal dynamics and anticipate future events; and \textbf{(4) \ulGeneralization} (\colorbox{green!10}{Gen.}), which measures whether the first three abilities remain reliable across diverse driving scenarios and data sources.
Table \ref{tab:benchmarks_comparison} provides a comparison between \dataset and the existing multi-view QA benchmarks for AD.

\noindent
Our contributions are: \textbf{(i)} We \textbf{introduce \dataset}, a spatiotemporal intelligence benchmark comprising 15.6k QA pairs across 20 tasks, curated from five AD datasets and spanning diverse weather, scene, and lighting conditions. \textbf{(ii)} We propose a \textbf{scalable benchmark construction pipeline} that combines a dynamic multi-relational scene graph with structured human verification to enforce cross-view and temporal constraints during QA generation. \textbf{(iii)} We \textbf{conduct an extensive evaluation} of 15 representative VLMs and show a 28.4-point human-model gap on \dataset-\textit{mini}, with \colorbox{red!10}{Const.} emerging as the dominant bottleneck. We further find that language-only prompting does not close this gap, while explicit spatial grounding consistently improves performance.
%Our analysis further demonstrates that pre-constructed spatial representations consistently improve model performance, highlighting a promising direction for advancing spatiotemporal reasoning in AD. 

\section{Related Works}
%\nghi{I split the related work into a third subsection on spatial/spatiotemporal intelligence in VLMs and folded in additional driving-QA, scene-graph, perspective-taking, and occlusion citations to better situate \dataset.}
% \hl{Hao: em check all citations if they are appropriated, for entire related work section}

\noindent
\textbf{VLMs in AD.}
Recent work has explored VLMs for end-to-end AD, including language-grounded perception, prediction, planning, and action generation~\cite{arai2025covla, li2024ego, hwang2024emma, wang2025omnidrive, xu2024drivegpt4, sima2024drivelm, tian2024drivevlm, zeng2025futuresightdrive, zhou2023vision, cui2023survey}. These models typically build on modern multimodal foundation models~\cite{bai2025qwen3, touvron2023llama, achiam2023gpt, wang2025internvl3, team2025kimi, li2024llava, wu2024deepseek, team2023gemini, vo2024henasy, truong2025directed}, while a parallel line of work uses generative world models~\cite{hu2023gaia, wang2024drivewm, wang2023drivedreamer} to capture multi-view scene dynamics. Most evaluations, however, still take place in relatively simple open-loop settings on nuScenes~\cite{tian2025nuscenes}, where large portions of the driving distribution correspond to low-complexity motion~\cite{li2024ego}, motivating spatial reasoning as a complementary evaluation lens for reliability and interpretability~\cite{wang2025omnidrive, sima2024drivelm, ma2024dolphins, huang2025robotron, nie2024reason2drive, zeng2025futuresightdrive}.

\noindent
\textbf{Spatial and Spatiotemporal Intelligence in VLMs.}
A growing body of work probes whether VLMs possess genuine spatial intelligence. General-purpose benchmarks evaluate static spatial relations~\cite{visualspatial, spatialmm, embspatial}, grounded and 3D-aware reasoning~\cite{spatialrgpt, spatialvlm, cai2025spatialbot, ma2025spatialllm, robospatial, erqa, liu2025spatialcot}, and broader capability suites~\cite{omnispatial, jia2025omnispatial, kang2025hssbench, huang2025spatialdise, stogiannidis2025mindgap}. Video-based benchmarks extend this to spatiotemporal reasoning~\cite{vsibench, lin2025mmsivideo, yang2025cambrians, grunde2021agqa, xiao2021nextqa}, and recent studies highlight specific failure modes such as egocentric bias and weak perspective-taking~\cite{prunty2026visuospatial, wang2026egocentricbias, wang2026allocentricperceiver, jang2026sympl}, occlusion-aware reasoning~\cite{pothiraj2025capture, liu2025obench}, and difficulty in maintaining cognitive maps from video~\cite{zhu2026mindoverspace, huang2025video2layout, zhao2025embodiedr}. \dataset{} complements these efforts by grounding spatiotemporal evaluation specifically in multi-view, multi-source driving scenes.

\noindent
\textbf{Benchmarks for Spatial Understanding in AD.}
Driving QA benchmarks span scene description, perception and prediction, driving advice, attention alignment, and counterfactual reasoning~\cite{deruyttere2019talk2car, gholami2025spatial, cao2024maplm, tian2025nuscenes, guo2025surds, wang2025omnidrive, ding2024holistic, sima2024drivelm, chen2025automated, wu2025towards, marcu2024lingoqa, kim2018textual, qian2024nuscenes, inoue2023nuscenesmqa, ishihara2025strideqa, jia2024bench2drive}, with several leveraging structured scene-graph representations of road actors and their relations~\cite{tian2020roadscenegraph, wang2023rs2g}. Despite this progress, most are limited by single-source data, static observations, single-view reasoning, or question forms answerable from local visual cues alone. \dataset{} complements rather than replaces these benchmarks: it spans multiple datasets and ego platforms, explicitly enforces cross-view and temporal constraints during question generation, and unifies scene construction, multi-view relational reasoning, and temporal reasoning under one taxonomy, making it a better fit for diagnosing whether a VLM can build and use a coherent scene-level representation in dynamic driving environments.

\begin{figure}[t]
  \centering
  \includegraphics[page=1, width=0.85\linewidth]{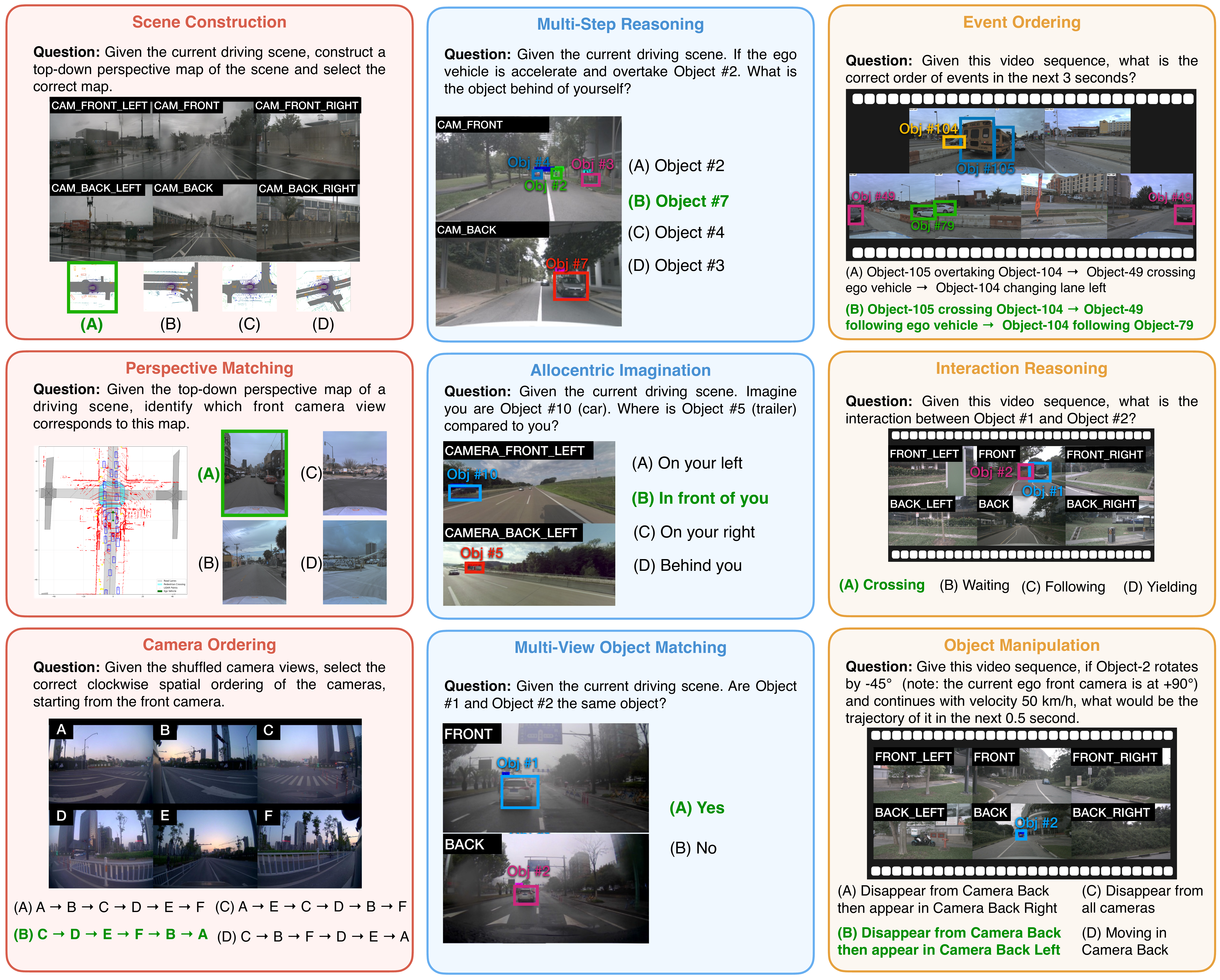}
  \caption{Representative question samples from \dataset across nine selected tasks (out of 20). Each cell shows a multiple-choice question with its visual input and answer options; correct answers are \textcolor{green!50!black}{\textbf{bold}}. Tasks are grouped by spatiotemporal ability: \colorbox{red!10}{Const.}, \colorbox{blue!10}{Unders.}, \colorbox{abilityTmp!10}{Reas.}
  %\textit{\protect\ulCognitiveScene}, \textit{\protect\ulMultiViewRel}, and \textit{\protect\ulTemporalReason}.
  } 
  \label{fig:task_examples}
\vspace{-5mm}
\end{figure}
\section{DriveSpatial}
%\subsection{Overview}
\dataset
is built on five large-scale AD datasets: nuScenes~\cite{caesar2020nuscenes}, Waymo~\cite{sun2020scalability}, TruckScenes~\cite{fent2024man}, AV2~\cite{wilson2023argoverse}, and ONCE~\cite{mao2021one}, spanning both car and truck platforms across diverse weather, scene layout, and time of day. \dataset independently probes three reasoning abilities and one robustness dimension: \colorbox{red!10}{Const.} asks whether a model can construct coherent scene representations from multi-view observations, \colorbox{blue!10}{Unders.} asks whether it can reason about spatial relationships across viewpoints, \colorbox{abilityTmp!10}{Reas.} asks whether it can leverage temporal context to infer dynamics and anticipate future events, and \colorbox{green!10}{Gen.} measures whether these abilities remain reliable across datasets and driving conditions.

\begin{figure}[t]
  \centering
  \includegraphics[page=1, width=.9\linewidth]{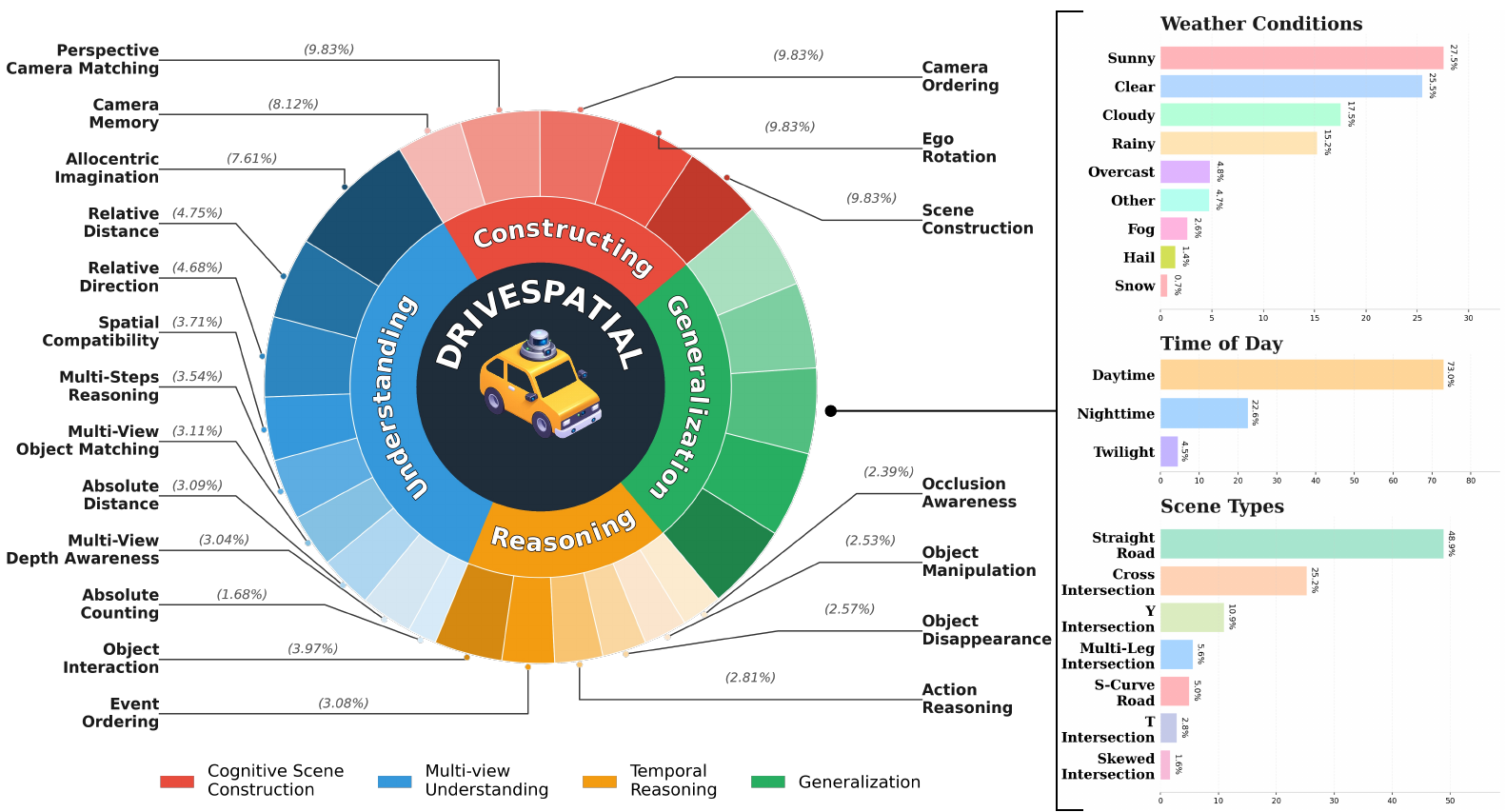}
  % \caption{\textbf{\dataset statistics.}  \vspace{-1em}}  \label{fig:data_statistic}
\caption{\textbf{\dataset statistics.} \textbf{(Left)} Sunburst view of the 20 tasks under abilities \colorbox{red!10}{Const.}, \colorbox{blue!10}{Unders.} and \colorbox{abilityTmp!10}{Reas.}. \textbf{(Right)} Scene-level diversity distribution (\colorbox{green!10}{Gen.}).}
\label{fig:data_statistic}
\vspace{-1mm}
\end{figure}

\noindent
\noindent\textbf{Task Taxonomy \& Statistics.} \dataset{} comprises 20 tasks and 15,670 QA pairs, designed to evaluate four core abilities underlying human spatial navigation: \colorbox{red!10}{Const.} (5 tasks), \colorbox{blue!10}{Unders.} (9 tasks), \colorbox{abilityTmp!10}{Reas.} (6 tasks), and \colorbox{green!10}{Gen.} Examples are shown in Figure~\ref{fig:task_examples}, and the distribution of tasks and datasets is summarized in Figure~\ref{fig:data_statistic}. Figure~\ref{fig:data_statistic} (Left) shows the distribution of QA tasks:  \colorbox{red!10}{Const.} (47.44\%), \colorbox{blue!10}{Unders.} (35.21\%), \colorbox{abilityTmp!10}{Reas.} (17.35\%). Figure~\ref{fig:data_statistic} (Right) shows \colorbox{green!10}{Gen.}, covering nine weather conditions, three periods of day and seven different scene types.
%, including challenging scenarios such as multi-leg intersections, snow, fog, and nighttime driving. 

% \textit{Scene Construction} (5 tasks), \textit{Multi-view Relational Understanding} (7 tasks), \textit{Temporal Reasoning} (6 tasks), and \textit{Generalization} (5 datasets).
% Each question contains either a single image or a sequence of frames (approximately 3 seconds), accompanied by a question and multiple candidate answers. 
% \NL{Move this to above} Scene Construction tasks evaluate the ability to build an internal representation of the driving layout. Multi-view Relational Understanding tasks require understanding perspectives from both the ego vehicle and other vehicles, as well as multi-step reasoning to arrive at the correct answer. Temporal Reasoning tasks provide approximately 3 seconds of driving video to assess memory, temporal reasoning, and the ability to anticipate future events. To ensure robustness and adaptability, questions are sampled across different datasets, weather conditions, traffic densities, scene types, and times of day. 

% \noindent
%  \textbf{Statistics.} \dataset{} comprises 20 tasks and 16,005 QA pairs derived from five AD benchmarks, designed to evaluate four core abilities underlying human spatial navigation. Figure~\ref{fig:data_statistic} (Left) shows the distribution of QA tasks: \ulSceneConstruction{} (45.77\%), \colorbox{blue!10}{Unders.} (36.34\%), \ulTemporalReason{} (17.89\%). Figure~\ref{fig:data_statistic} (Right) shows that \dataset covers seven different map types, nine weather conditions,  and three periods of the day, including challenging scenarios such as multi-leg intersections, snow, fog, and nighttime driving. 

\subsection{Benchmark Construction Pipeline}
\begin{figure}[t]
  \centering
  \includegraphics[page=1, width=.9\linewidth]{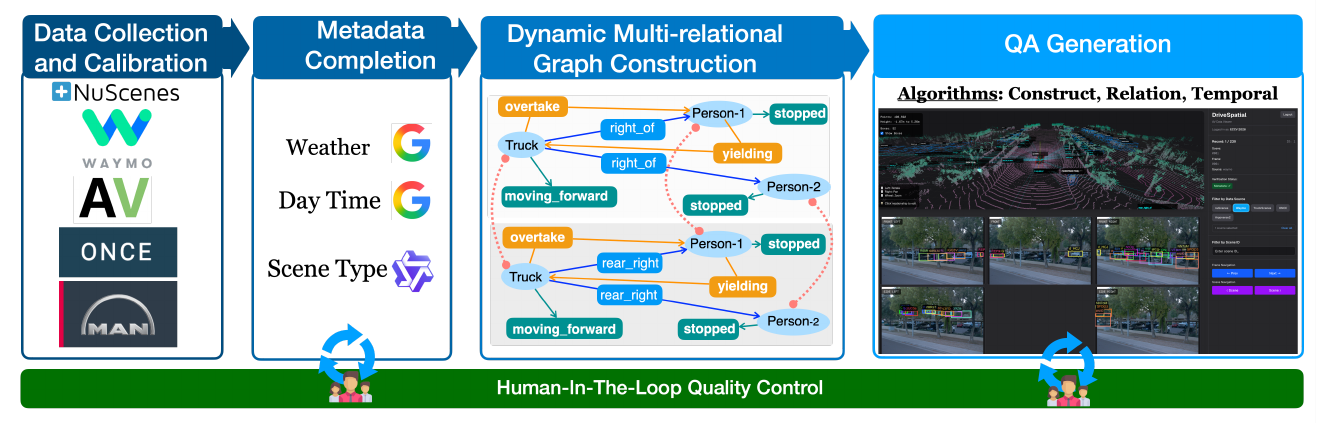}
  \caption{\textbf{\dataset construction pipeline.} (1) standardize five AV datasets into a unified schema; (2) complete scene-level metadata; (3) construct a dynamic multi-relational graph; and (4) apply 20 rule-based algorithms to generate QA pairs. To ensure quality, human-in-the-loop is applied. \vspace{-1em}}
  \label{fig:data_creation}
  \vspace{-8mm}
\end{figure}

\dataset construction pipeline includes four stages, as in Figure~\ref{fig:data_creation}.
Human-in-the-loop verification is applied at both the metadata and QA generation stages to ensure annotation quality. Full details are provided in Appendix \ref{app:benchmark_construction}. 

\noindent\textbf{Data Collection and Calibration.} We standardize five public AD validation sets ~\cite{sun2020scalability,caesar2020nuscenes,wilson2023argoverse,fent2024man,mao2021one} into a unified interface. Because these datasets do not share the same coordinate convention, we adopt nuScenes as the reference system and align all other datasets.
%by applying a rotation offset $\Delta\alpha$ specific to each source. 
%All sources are then parsed into a unified database schema, enabling downstream stages to operate identically regardless of the data origin.

% \noindent\textbf{Metadata Completion.} Although the five source datasets collectively span a wide range of driving conditions, their scene-level annotations are incomplete: weather labels are available in only three datasets, time-of-day in four, and scene type in none. Since subsequent stages, namely graph-based question generation and cross-condition generalization analysis, depend on these attributes, we systematically complete them. We infer \textit{weather} and \textit{time of day} using SigLIP-2~\cite{tschannen2025siglip} through image\textendash text similarity matching on uniformly sampled keyframes, and classify \textit{scene type} by rendering the top-down map and querying Qwen3-VL-8B~\cite{yang2025qwen3} with a structured layout prompt. All inferred labels are verified by domain experts before being synchronized into the unified database. This completed metadata enables the QA generation stage to sample questions across diverse conditions and supports the generalization analysis reported in Section~\ref{sec:experiments}.

\noindent\textbf{Metadata Completion.} We complete missing scene-level annotations required for graph-based QA generation and generalization analysis. We infer \textit{weather} and \textit{time of day} via SigLIP-2~\cite{tschannen2025siglip} image\textendash text similarity on sampled keyframes, and classify \textit{scene type} by querying Qwen3-VL-8B~\cite{yang2025qwen3} on rendered top-down maps. All inferred labels are then verified by domain experts.

\noindent\textbf{Dynamic Multi-relational Graph Construction.}
%\nghi{I made the multi-view claim explicit here by stating the camera-visibility constraints used during QA generation, so the benchmark now explains how cross-view reasoning is enforced instead of only asserting it.}
We construct a dynamic multi-relational graph $\mathcal{G}$ that encodes the spatial structure, interactions, actions, and temporal evolution of all objects in a scene: $\mathcal{G} = \left(\bigcup_{t=1}^{T} \mathcal{V}^t,\; \mathcal{E}_R \cup \mathcal{E}_I \cup \mathcal{E}_A \cup \mathcal{E}_T\right),$
% \begin{equation}
% \mathcal{G} = \left(\bigcup_{t=1}^{T} \mathcal{V}^t,\; \mathcal{E}_R \cup \mathcal{E}_I \cup \mathcal{E}_A \cup \mathcal{E}_T\right),
% \label{eq:graph_definition}
% \end{equation}
where $\mathcal{V}^t = \{v_i^t\}_{i=1}^{N_t}$ is the set of objects at timestep $t$, each characterized by 3D position, direction, semantic category, and visible camera views. The four edge types capture \textit{spatial} relationships $\mathcal{E}_R$ (\textcolor{blue}{$\rightarrow$}, e.g., \textit{left\_of}), pairwise \textit{interactions} $\mathcal{E}_I$ (\textcolor{orange}{$\rightarrow$}, e.g., \textit{overtaking}), per-object \textit{actions} $\mathcal{E}_A$ (\textcolor{teal}{$\circlearrowright$}, e.g., \textit{stopped}) as self-loops, and \textit{temporal} links $\mathcal{E}_T$ (\textcolor{Salmon}{$\bullet$\hdashrule{1cm}{2pt}{.5mm}\hspace{-.15em}$\bullet$}) between nodes sharing a track ID across consecutive frames. For tasks intended to require cross-view reasoning, we impose explicit camera constraints: $\lvert \mathrm{cam}(v_i^t)\rvert \geq 2$ for cross-view identity queries, and $\mathrm{cam}(v_i^t) \cap \mathrm{cam}(v_j^t) = \emptyset$ for pairwise relation queries. These constraints prevent the answer from being recovered from a single camera alone.

% \noindent\textbf{QA Generation.} Using the constructed graph $\mathcal{G}$ and completed metadata as input, we develop 18 rule-based algorithms organized into three families: \textit{Construct}, \textit{Relation}, and \textit{Temporal}, corresponding to the three task categories. Each algorithm traverses the graph to extract the relevant nodes and edges for a given task, then applies predefined question templates to produce QA pairs. For example, allocentric imagination queries sample a pair of objects $(v_i^t, v_j^t)$ from different camera views connected by a spatial edge in $\mathcal{E}_R$ and ask the model to infer the relative position of one object from the other's viewpoint, while interaction reasoning queries extract temporal sequences of interaction edges from $\mathcal{E}_I$ between objects observed across different views and ask the model to identify the pairwise behavior. To ensure diversity, the sampling strategy conditions on scene-level metadata such as weather, scene type, and time of day, balancing the distribution across driving conditions. The complete set of algorithms and question templates is provided in Appendix~\textcolor{blue}{A.4}.

\noindent\textbf{QA Generation.} Using $\mathcal{G}$ and the completed metadata, we develop 20 rule-based %algorithms organized into three families, 
QA generators in three families: \textit{Construct}, \textit{Relation}, and \textit{Temporal}, corresponding to the three task categories. Each algorithm traverses the graph to extract the relevant nodes and edges, then instantiates predefined question templates to produce QA pairs. 

\noindent\textbf{Human-In-The-Loop Quality Control.} In parallel, annotators verify scene-level metadata through a web-based interface (Figure~\ref{fig:data_creation} Right) and validate each QA pair via a separate interface for answer correctness, option uniqueness, question plausibility, and object visibility. In total, this verification stage required approximately 160 annotator-hours, with 76.86\% of generated questions passing all quality criteria; the remainder were either revised or discarded.
% \noindent\textbf{Human-In-The-Loop Quality Control.} Annotators verify each QA pair for answer correctness, option uniqueness, question plausibility, and object visibility. %Annotators evaluate each generated QA pair against four criteria: (1) answer correctness, (2) option uniqueness, (3) question plausibility, and (4) object visibility. %Across the benchmark, annotators dedicated over 80 hours to this stage, yielding a 76.86\% acceptance rate and demonstrating that graph-based generation produces high-quality QA pairs at scale under human oversight. 

\section{Benchmark Evaluation}
\label{sec:experiments}
\subsection{Evaluation Setup}
\noindent
\textbf{Benchmark models.} We evaluate 15 VLMs across diverse model families, scales, and training paradigms, grouped into three categories: proprietary models (\textit{Proprietary})~\cite{team2024gemini,hurst2024gpt,singh2025openai}, general-purpose open-source models (\textit{Generalists}), including both image-based~\cite{li2024llava,wu2024deepseek,kamath2025gemma,zhu2025internvl3,wang2025internvl3,bai2025qwen3} and video-based architectures~\cite{zhang2024long,zhang2024llava}, and domain-specialized models (\textit{Specialists}) for autonomous driving and spatial reasoning~\cite{huang2025robotron,gholami2025spatial} or 3D embodied tasks~\cite{cai2025scaling,batra2025spatialthinker}.
%We evaluate 15 VLMs spanning diverse model families, parameter scales, and training paradigms. Our evaluation covers three categories. First, we include proprietary models (Commercialists) ~\cite{team2024gemini,hurst2024gpt,singh2025openai}. Second, for general-purpose open-source models (Generalists), we evaluate both image-based ~\cite{li2024llava,wu2024deepseek,kamath2025gemma,zhu2025internvl3,wang2025internvl3,bai2025qwen3} and video-based architectures ~\cite{zhang2024long,zhang2024llava}. Third, for domain-specialized models (Specialists), we consider two groups: state-of-the-art models for end-to-end driving and spatial reasoning in autonomous vehicles ~\cite{huang2025robotron,gholami2025spatial}, as well as models tailored to 3D embodied tasks ~\cite{cai2025scaling,batra2025spatialthinker}.

\noindent
\textbf{Evaluation Tasks and Metrics.} 
%To quantitatively evaluate model performance on the \dataset dataset, we adopt metrics tailored to each task type.
For multiple-choice answer (MCA) tasks, we follow %standard practice in 
prior MCA benchmarks~\cite{fu2025video, hendrycks2020measuring, yue2024mmmu, yang2025thinking} and report exact-match accuracy. 
%accuracy based on exact matching as the primary metric. 
For open-ended tasks requiring absolute numeric predictions (e.g., distances, counts, or spatial measurements), we report the root mean squared error (RMSE) against ground truth.
%between predicted and ground-truth values to capture deviations in continuous-valued outputs. 
Additionally, we report Cohen's $\kappa$ agreement~\cite{cohen1960coefficient} to assess the alignment between human annotators and VLM predictions.

\noindent
\textbf{Baselines.} To contextualize model performance, we establish three baselines: two chance-level baselines including \textit{Random} (uniform answer selection) and \textit{Frequency} (the most frequent answer), and a \textit{Human-Level} on \dataset-\textit{mini}, a stratified subset of 1,000 questions (50 per task). Human annotators answer each question independently. Full implementation details, including frame sampling, camera ordering, context limits, and prompt templates, are provided in Appendix~\ref{supp:evaluate_setup}.
%We first report two chance-level baselines: \textit{Random}, based on uniform answer selection, and \textit{Frequency}, which always chooses the most frequent answer for each task. We also establish a \textit{Human-Level} baseline on DriveSpatial-\textit{mini}, a stratified subset of 1,000 questions (50 per task). Human annotators answer each question independently and are evaluated with the same metrics. Further details are provided in Appendix \textcolor{blue}{B}.

\begin{figure*}[t]
  \centering
  % LEFT — Table
  \begin{minipage}[t]{0.6\textwidth}
    \vspace{0pt}\centering
    \fontsize{6pt}{6.5pt}\selectfont
    \setlength{\tabcolsep}{2pt}
    \renewcommand{\arraystretch}{1.2}
    \resizebox{\linewidth}{!}{%
    \begin{tabular}{l|c|c|c|cc|c}
      \toprule
      \multirow{2}{*}{\textbf{Methods}} &  \multirow{2}{*}{\textbf{Rank}} &  \multirow{2}{*}{\textbf{Avg.}} &
      \multicolumn{1}{c}{\cellcolor{red!10}Const.} &
      \multicolumn{2}{c|}{\cellcolor{blue!10}Unders.} &
      \multicolumn{1}{c}{\cellcolor{abilityTmp!10}Reas.} \\
      \cmidrule(lr){4-4} \cmidrule(lr){5-6} \cmidrule(lr){7-7}
      & & &
      \colorbox{red!10}{Acc}$\uparrow$ &
      \colorbox{blue!10}{Acc}$\uparrow$ &
      \colorbox{blue!10}{RMSE}$\downarrow$ &
      \colorbox{abilityTmp!10}{Acc}$\uparrow$ \\
      \midrule
      \rowcolor{gray!15}
      \multicolumn{7}{l}{\textit{Baselines}} \\
      Random & 13 & 26.33 & 25.37 & 28.24 & -- & 25.39 \\
      Frequency & 8 & 32.45 & 32.94 & 33.89 & -- & 30.51 \\
      Human\textsuperscript{$\ddagger$} & 1 & 83.39 & 86.20 & 85.62 & 10.62 & 88.96 \\
      \midrule
      \rowcolor{gray!15}
      \multicolumn{7}{l}{\textit{Proprietary}}  \\
      GPT-4o\textsuperscript{$\ddagger$}\cite{hurst2024gpt} & 3 & 51.37 & 48.22 & 59.84 & 12.71 & 58.76 \\
      GPT-5\textsuperscript{$\ddagger$} \cite{singh2025openai}& 2 & 54.98 & 55.20 & 62.45 & 10.41 & 57.69 \\
      Gemini-2 Pro\textsuperscript{$\ddagger$} \cite{team2024gemini} & 4 & 47.26 & 44.09 & 58.51 & 13.20 & 52.38 \\
      \midrule
      \rowcolor{gray!15}
      \multicolumn{7}{l}{\textit{Generalist}} \\
      LLaVA-Onevision-7B \cite{li2024llava}& 10 & 28.65 & 33.73 & 27.64 & 15.76 & 40.34 \\
      DeepSeek-VL2-Small \cite{wu2024deepseek} & 14 & 24.88 & 23.92 & 26.60 & 15.15 & 39.26 \\
      Gemma-3-12B-it \cite{kamath2025gemma} & 7 & 35.25 & 35.55 & 44.05 & 14.65 & 40.80 \\
      InternVL-3.5 8B \cite{zhu2025internvl3} & 6 & 36.73 & 39.00 & 44.15 & 14.26 & 41.31 \\
      InternVL-3 8B \cite{wang2025internvl3} & 16 & 24.20 & 30.63 & 27.26 & 15.95 & 30.65 \\
      Qwen3-VL 8B \cite{bai2025qwen3}& 5 & 42.24 & 42.76 & 50.62 & 14.32 & 47.66 \\
      LongVA-7B \cite{zhang2024long}& 12 & 25.91 & 26.84 & 35.90 & 20.26 & 35.25 \\
      LLaVA-Video-7B \cite{zhang2024llava} & 13 & 25.24 & 26.94 & 27.57 & 15.68 & 36.89\\
      \midrule
      \rowcolor{gray!15}
      \multicolumn{7}{l}{\textit{Specialist}} \\
      RoboTron-Drive \cite{huang2025robotron} & 18 & 23.14 & 24.31 & 21.03 & 15.49 & 39.56 \\
      Ego3D \cite{gholami2025spatial} & 11 & 26.48 & 30.21 & 30.67 & 15.95 & 34.53 \\
      SpaceThinker \cite{batra2025spatialthinker} & 17 & 24.53 & 27.16 & 31.74 & 16.96 & 31.65 \\
      SenseNova-SI \cite{cai2025scaling} & 9 & 30.54 & 37.91 & 36.25 & 15.25 & 32.73 \\
      \bottomrule
    \end{tabular}
    }
  \end{minipage}%
  \hspace{0.005\textwidth}%
  % RIGHT — Chart on top, caption on bottom
  \begin{minipage}[t]{0.39\textwidth}
    \vspace{-2mm}
    \centering
    \includegraphics[width=.97\linewidth]{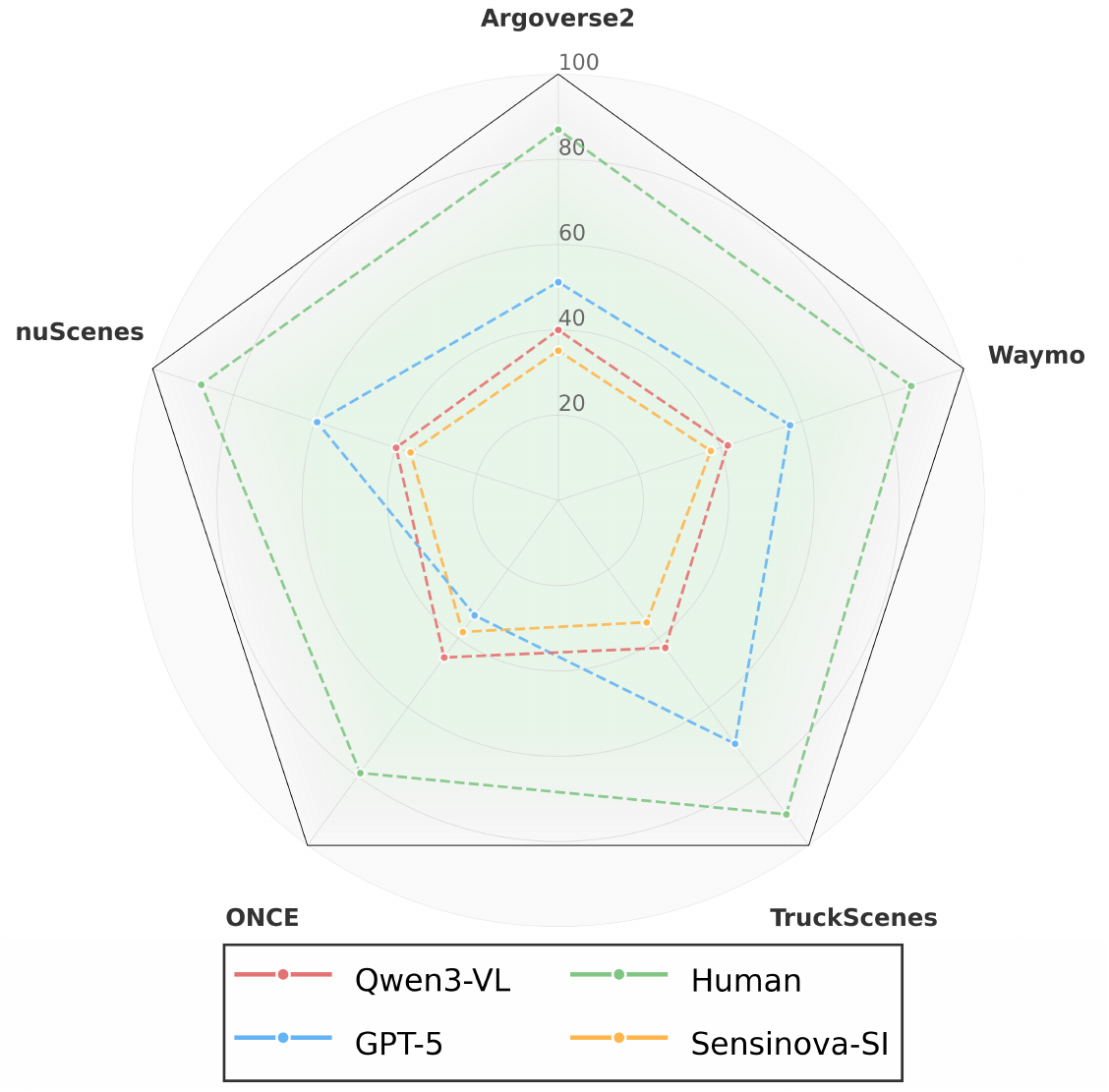}
    \captionsetup{type=table, justification=justified, singlelinecheck=false, font=small}
    \vspace{3mm}
    \caption{Evaluation on \dataset{}. \textbf{Left:} Overall benchmarking results on \colorbox{red!10}{Const.}, \colorbox{blue!10}{Unders.}, and \colorbox{abilityTmp!10}{Reas.}. \textsuperscript{$\ddagger$} indicates models evaluated on \dataset-\textit{mini}. Avg.\ is computed as $\frac{(\text{\colorbox{red!10}{Acc}} + \text{\colorbox{blue!10}{Acc}} - \text{\colorbox{blue!10}{RMSE}} + \text{\colorbox{abilityTmp!10}{Acc}}}{3}$. \textbf{Right:} Results of the best-performing model in each group across five data sources (\colorbox{green!10}{Gen.}).}
    \label{fig:main_result}
  \end{minipage}
  \vspace{-0.4cm}
\end{figure*}

\subsection{Overall Benchmark Results} 
%\nghi{I reorganized this results section so the main message is immediate: the large human-model gap, scene construction as the main bottleneck, and BEV grounding as a diagnostic aid rather than a deployable module.}
Table~\ref{fig:main_result} summarizes the performance on \dataset{} and highlights findings:
\textbf{(i) A pronounced human-model gap persists, and scene construction is the main bottleneck.} On \dataset-\textit{mini}, humans (83.39\%), outperforms GPT-5 (54.98\%) by 28.41 points and the best open-source Qwen3-VL 8B (42.24\%), by more than 41 points.
%The gap is driven primarily by \colorbox{red!10}{Const.} as we obverse a model perform better on \colorbox{red!10}{Const.} tend to perform better in other two tasks. By contrast,
Human on numeric \colorbox{blue!10}{Unders.} tasks (10.62) is close to GPT-5 (10.41), suggesting that current VLMs can sometimes estimate geometry, but still struggle to build a coherent multi-view scene representation.
\textbf{(ii) Generalists outperform Specialists on \dataset{}, reversing the trend reported on earlier AD benchmarks.} Specialist driving models such as RoboTron-Drive and Ego3D fall to 23.14\% and 26.48\%, both below the \textit{Frequency} baseline (32.45\%). A similar pattern holds for 3D-embodied specialists such as SpaceThinker and SenseNova-SI. This suggests that training centered on single-view perception and short templated QA transfers poorly to our cross-view, longer-horizon, and compositional evaluation setting.
\textbf{(iii) Per-source performance varies substantially across model families.} The per-source breakdown in Fig.~\ref{fig:main_result} (right) shows that human performance is relatively stable across all five data sources, whereas VLM performance fluctuates more strongly. GPT-5 performs well on nuScenes, Waymo, AV2, and TruckScenes but drops on ONCE, where Qwen3-VL 8B and SenseNova-SI perform better. Since ONCE differs geographically and operationally from the other sources, this pattern suggests that current VLMs remain sensitive to source shift.

\subsection{Ablation Studies}
\begin{figure}[h]
  \centering
  \includegraphics[page=1, width=\linewidth]{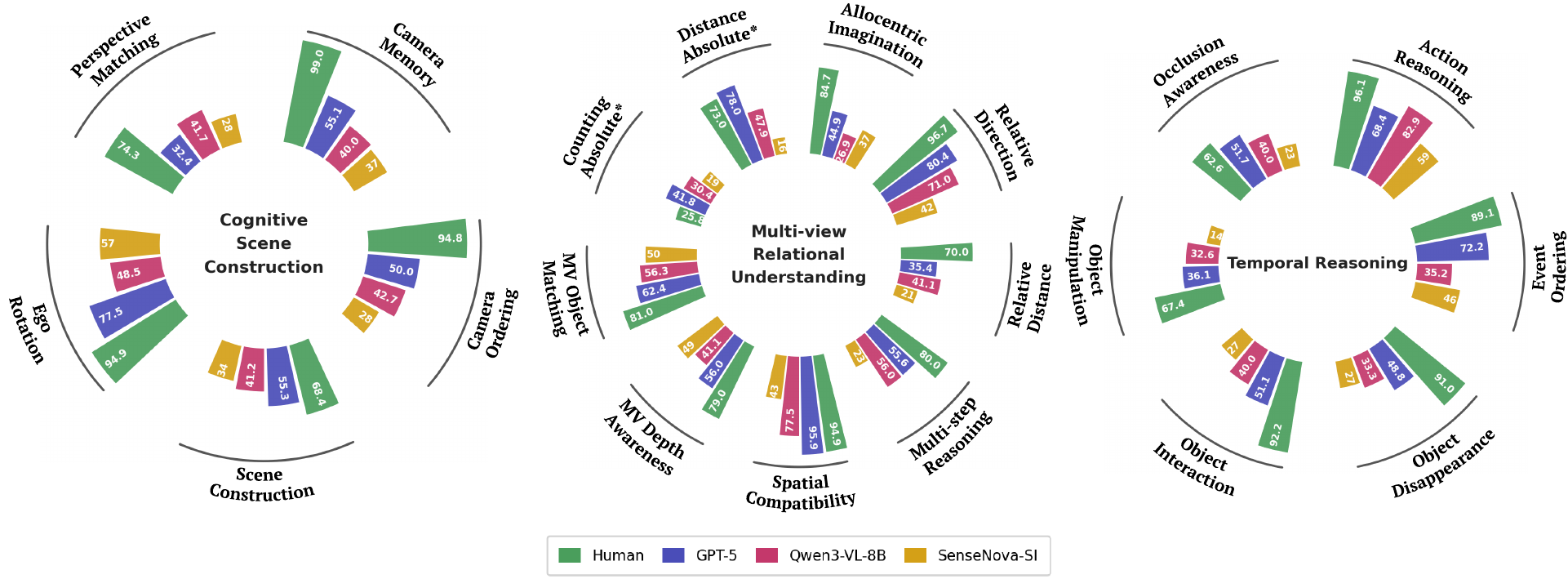}
  % \caption{Task-level breakdown of VLMs and comparison with human performance. $\ast$ denotes task evaluated with RMSE metric. For visualization we report $\ast$ with this formulation: $\mathrm{Score} = \frac{\hat{y} - y}{\hat{y}} \times 100$, where $\hat{y}$ is the reference tolerance and $y$ is the predicted error,  $\hat{y}$ is seted as 25 for Counting Absolute and 10 for Distance Absolute, respectively. \textit{Best view in zoom}.}
  \caption{Per-task comparison against human performance. (left $\rightarrow$ right: \colorbox{red!10}{Const.}, \colorbox{blue!10}{Unders.}, \colorbox{abilityTmp!10}{Reas.}) $^\ast$: \colorbox{blue!10}{RMSE}-evaluated tasks, rescaled as $(\hat{y} - y)/\hat{y} \times 100$ for visualization, where $y$ is the predicted error and $\hat{y}$ is a task-specific tolerance (10 for \textit{Counting Absolute}, 25 for \textit{Distance Absolute}). }
  \label{fig:perf_breakdown}
\end{figure}

\noindent
\textbf{Ability-Level and Task-Level Analysis}
Figure~\ref{fig:perf_breakdown} reveals a consistent trend across all evaluated models: \colorbox{red!10}{Const.} is the weakest ability group, and models that perform better on this task tend to also perform better on other two tasks%\textit{\ulMultiViewRel} and \textit{\ulTemporalReason} tasks
, suggesting that \colorbox{red!10}{Const.} is an upstream bottleneck.
%. This pattern suggests that scene construction acts as an upstream bottleneck for downstream spatial abilities.
In \colorbox{red!10}{Const.}, the largest human-to-model gaps appear on tasks such as \texttt{Camera Memory} (99.0 vs.\ 55.1 for GPT-5)
%the human-to-model gap is largest on construction-heavy tasks such as \texttt{Camera Memory} (99.0 vs.\ 55.1 for GPT-5, a 43.9-point gap) 
and \texttt{Camera Ordering} (94.8 vs.\ 50.0), indicating that current VLMs struggle to build and maintain coherent internal 3D representations from multi-view observations.
In \colorbox{blue!10}{Unders.}, the gap is smaller on geometry-centric tasks e.g., \texttt{Relative Direction} (96.7 vs.\ 80.4) and \texttt{Spatial Compatibility} (94.9 vs.\ 95.9), but remains large on perspective-dependent tasks e.g., \texttt{Allocentric Imagination} (84.7 vs.\ 44.9), highlighting the difficulty of egocentric-to-allocentric reasoning while explaining why prior benchmarks focusing on object-pair relations, tend to report higher model accuracy. 
%, where GPT-5 matches human performance), indicating that VLMs can partially leverage geometric priors learned during large-scale pretraining, particularly for depth- and direction-related reasoning. The relatively strong performance on these relational tasks also helps explain why prior benchmarks, which focus primarily on object-pair relations, tend to report higher model accuracy. In contrast, tasks that require perspective transformation, such as \texttt{Allocentric Imagination} (84.7 vs.\ 44.9), remain substantially harder, reflecting the persistent difficulty of egocentric-to-allocentric reasoning.
In \colorbox{abilityTmp!10}{Reas.}, models perform better on simpler dynamics, e.g., \texttt{Action Reasoning} (96.1 vs.\ 68.4 for GPT-5) and \texttt{Object Disappearance} (91.0 vs.\ 48.8), but remains weak on multi-step temporal tasks e.g., \texttt{Object Manipulation} (67.4 vs.\ 36.1) and \texttt{Occlusion Awareness} (62.6 vs.\ 51.7), indicating that temporally compositional reasoning remains a major challenge.
%which require tracking entities and reasoning over compositional state changes across time. This pattern suggests that temporally compositional reasoning is a key remaining challenge. 
Additional analysis is in Appendix~\ref{supp:furture_analysis}.

\begin{figure}[h]
  \centering
  \vspace{-1em}
  \includegraphics[page=1, width=\linewidth]{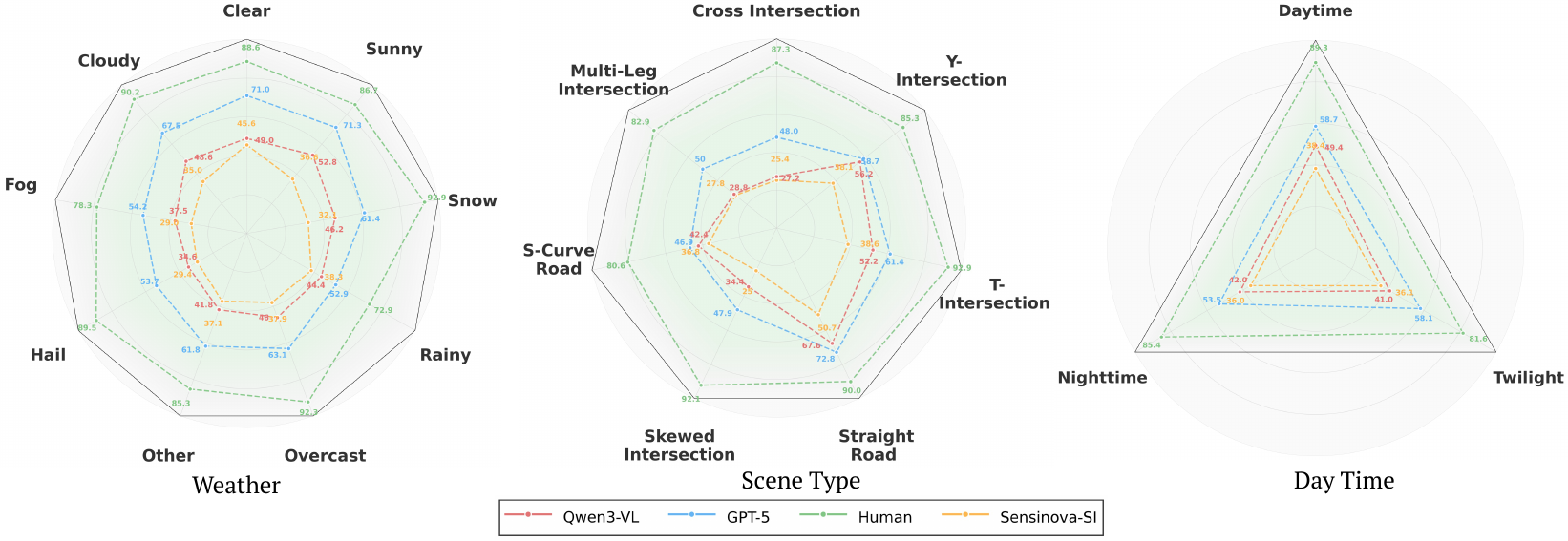}
  \caption{Breakdown of VLM performance for testing \colorbox{green!10}{Gen.} across diverse conditions.\vspace{-2em}}
  \label{fig:driving_codition}
\end{figure}

% \noindent 
% \textbf{Robustness to driving conditions.}
\noindent
\textbf{Generalization Across Sources and Driving Conditions}
Figure~\ref{fig:driving_codition} analyzes performance across weather, scene type, and time of day.
%Figure~\ref{fig:driving_codition} presents the performance analysis under different weather, scene layouts, and times of day.
%
Under adverse weather, both humans and models degrade, but human accuracy remains relatively stable (around 80\%), whereas VLMs drop sharply, e.g., GPT5 falls to 52.9\% in \texttt{rainy} scenes and 53.7\% in \texttt{hail}. 
%With respect to \textit{weather}, a clear trend emerges: both models and humans experience performance degradation in \texttt{rainy} scenes due to reduced visual clarity. Nevertheless, humans maintain relatively stable performance across conditions (around 80\%). While GPT-4o drops to 32.9\% and 47.4\% accuracy under \texttt{rainy} and \texttt{hail} conditions, respectively. 
%
Across scene types, humans remain robust, while VLMs struggle in complex layouts such as \texttt{cross intersections}, \texttt{multi-leg intersections}, and \texttt{skewed intersections}, which require accurate scene construction.
%In the \textit{scene-type} breakdown, human performance remains consistently high across different environments. In contrast, complex scenes such as \texttt{cross intersections}, \texttt{multi-leg intersections}, and \texttt{skewed intersections} pose challenges for VLMs. These scenarios require accurate scene construction to correctly infer the road layout, while the typically higher traffic density further complicates scene understanding and reasoning. 
%
Across times of day, performance is relatively balanced, though \texttt{nighttime} is the overall challenging condition: GPT-5, Qwen3-VL, and Sensinova-SI reach only 53.5\%, 42.0\%, and 36.0\%, respectively. 
These findings suggest that visual degradation is a major factor limiting VLMs reliability, motivating the integration of complementary modalities such as LiDAR.

%In the \text{time-of-day} analysis, we observe a clear performance degradation on questions related to \texttt{nighttime} scenes: GPT-4o, Qwen3-VL, and Ego3D achieve only 27.0\%, 18.0\%, and 10.5\% accuracy, respectively. This trend is consistent with the results observed under \texttt{rainy} conditions, where visual quality is degraded and reasoning becomes more difficult. These findings suggest that visual degradation plays a critical role in the reliability of VLMs. To improve robustness, future approaches may need to incorporate vision-independent modalities, such as LiDAR, to complement visual inputs when integrating with VLM-based systems.

% \noindent \textbf{How aligned are humans and VLMs?}
\noindent
\textbf{Human--Model Agreement}
Table~\ref{tab:kappa_agreement} reports the agreement between human annotations and VLM predictions measured by Cohen's $\kappa$. 
\begin{wraptable}{r}{0.4\textwidth}
\vspace{-1.2em}
\centering
\caption{Cohen's $\kappa$ agreement between human and model. F., S., and M.\: \textit{Fair}, \textit{Slight}, and \textit{Moderate} agreement levels.}
\label{tab:kappa_agreement}
\setlength{\tabcolsep}{3pt}
\renewcommand{\arraystretch}{1.1}
\resizebox{0.4\textwidth}{!}{%
\begin{tabular}{l|c|c|c|c}
\toprule
 & \textbf{GPT-5} & \textbf{Qwen3-VL} & \textbf{Ego-3D} & \textbf{Sensinova-SI} \\
\midrule
\cellcolor{red!10}Const.    & 0.35 (F.) & 0.09 (S.) & 0.11 (S.) & 0.08 (F.) \\
\cellcolor{blue!10}Unders.  & 0.43 (M.) & 0.33 (F.) & 0.10 (S.) & 0.09 (S.) \\
\cellcolor{abilityTmp!10}Reas.    & 0.36 (F.) & 0.24 (F.) & 0.15 (S.) & 0.08 (F.) \\
\bottomrule
\end{tabular}}
\vspace{-1em}
\end{wraptable}
GPT-5 achieves the highest agreement across all three task groups, but its scores remain modest: 0.35 (\textit{Fair}) on \colorbox{red!10}{Const.}, 0.43 (\textit{Moderate}) on \colorbox{blue!10}{Unders.}, and 0.36 (\textit{Fair}) on \colorbox{abilityTmp!10}{Reas.}.
%Among the evaluated models, GPT-5 shows the highest agreement with humans, outperforming Qwen3-VL, Ego-3D, and Sensinova-SI across all three task groups. Nevertheless, the Cohen's $\kappa$ values for GPT-5 remain relatively modest, reaching only 0.35 (\textit{Fair}), 0.43 (\textit{Moderate}), and 0.36 (\textit{Fair}) on construction, understanding, and reasoning tasks, respectively. 
The remaining models fall largely within the \textit{Slight}-to-\textit{Fair} range, with Ego-3D and Sensinova-SI staying below 0.15 throughout. These results indicate that even the strongest commercial models remain noticeably misaligned with human judgments, 
%still exhibit a noticeable gap in decision alignment with human judgments
, while open-source and domain-specific models show substantially weaker alignment.

%\noindent \textbf{Can spatiotemporal intelligence benefit from prompting techniques?} 
\noindent
\textbf{Diagnostic Studies}

\noindent \textbf{A. Impact of Linguistic Prompting.} We first examine whether standard prompting strategies improve performance on \dataset{}. Following prior work~\cite{huang2022language,jimenez2023swe,suris2023vipergpt,yang2025thinking,li2024advancing}, we evaluate VLMs from distinct families: Qwen3-VL, Gemma-3 and Ego3D. Specifically, we consider \textit{Chain-of-Thought} (CoT)~\cite{kojima2022large,wei2022chain}, \textit{Tree-of-Thoughts} (ToT)~\cite{yao2023tree}, and \textit{Self-Consistency} (SC)~\cite{wang2022self}. As shown in Table~\ref{tb:all-ablation} (Left) (Left), none of these methods improves over the zero-shot (ZS) baseline under our evaluation setup; instead, all consistently degrade performance across the three models. The drop is largest for CoT, which reduces average accuracy by $1.63\%$ , $0.53\%$, and $2.2\%$ for Qwen3-VL, Gemma-3, and Ego3D, respectively. ToT and SC are less harmful, but still underperform ZS. This pattern suggests that, for \dataset{}, textual reasoning scaffolds do not compensate for missing geometric grounding.
%\vspace{-.5em}
\begin{table*}[h]
\centering
%\scriptsize
\caption{\vspace{-.5em} Ablation of Linguistic Prompting; Geometric Grounding; Number of Views (\textbf{Left} $\rightarrow$ \textbf{Right}).}
\label{tb:all-ablation}
\renewcommand{\arraystretch}{0.9}
\setlength{\tabcolsep}{3pt}

% ----- LEFT: Prompting Techniques (full height) -----
\begin{minipage}[t]{0.34\textwidth}
  \vspace{0pt}
  \centering
  \resizebox{\linewidth}{!}{%
  \begin{tabular}{c|l|c|c|cc|c}
    \toprule
     & \multirow{3}{*}{\textbf{Model}} & \multirow{3}{*}{\textbf{Avg.}} &
    \multicolumn{1}{c}{\cellcolor{red!10}\textbf{Const.}} &
    \multicolumn{2}{c|}{\cellcolor{blue!10}\textbf{Unders.}} &
    \multicolumn{1}{c}{\cellcolor{abilityTmp!10}\textbf{Reas.}} \\
    \cmidrule(lr){4-4} \cmidrule(lr){5-6} \cmidrule(lr){7-7}
     & & &
     \colorbox{red!10}{Acc}$\uparrow$ &
     \colorbox{blue!10}{Acc}$\uparrow$ &
     \colorbox{blue!10}{RMSE}$\downarrow$ &
     \colorbox{abilityTmp!10}{Acc}$\uparrow$ \\
    \midrule
    \multirow{3}{*}{\rotatebox[origin=c]{90}{\makecell[c]{ZS}}}
     & Qwen3-VL & 42.24 & 42.76 & 50.62 & 14.32 & 47.66 \\
     & Gemma-3  & 35.25 & 35.55 & 44.05 & 14.65 & 40.80 \\
     & Ego3D    & 26.48 & 30.21 & 30.67 & 15.95 & 34.53 \\
    \addlinespace[3pt]\midrule
    \multirow{3}{*}{\rotatebox[origin=c]{90}{\makecell[c]{CoT}}}
     & Qwen3-VL & 40.61 & 40.10 & 50.04 & 14.52 & 46.20 \\
     & Gemma-3  & 34.72 & 35.59 & 44.04 & 14.25 & 38.78 \\
     & Ego3D    & 24.28 & 29.40 & 26.60 & 15.80 & 32.64 \\
    \addlinespace[3pt]\midrule
    \multirow{3}{*}{\rotatebox[origin=c]{90}{\makecell[c]{ToT}}}
     & Qwen3-VL & 40.58 & 40.02 & 47.02 & 13.69 & 48.38 \\
     & Gemma-3  & 34.99 & 36.02 & 41.42 & 13.27 & 40.80 \\
     & Ego3D    & 25.55 & 31.53 & 26.99 & 14.92 & 33.06 \\
    \addlinespace[3pt]\midrule
    \multirow{3}{*}{\rotatebox[origin=c]{90}{\makecell[c]{SC}}}
     & Qwen3-VL & 41.23 & 41.20 & 50.10 & 13.68 & 46.08 \\
     & Gemma-3  & 35.14 & 35.62 & 44.02 & 14.01 & 39.80 \\
     & Ego3D    & 25.12 & 30.46 & 26.48 & 15.02 & 33.43 \\
    \bottomrule
  \end{tabular}%
  }
\end{minipage}%
\hfill
% ----- RIGHT: caption on top, table on bottom -----
\begin{minipage}[t]{0.3\textwidth}
  \vspace{0pt}
  % --- caption block at the top ---
  % \captionsetup{type=table}
  % \caption{Comparison of the effectiveness of: (\textbf{Left}) -- linguistic prompting and (\textbf{Right}) -- pre-constructed BEV map prompting techniques.} %\textbf{ZS}: Zero-shot, \textbf{CoT}: Chain-of-Thought, \textbf{ToT}: Tree-of-Thoughts, \textbf{SC}: Self-Consistency.}
%  \label{tab:cot_method}

  \vfill

  % --- table pinned to the bottom ---
  \centering
  \resizebox{\linewidth}{!}{%
  \begin{tabular}{c|l|c|cc|c}
    \toprule
     {\textbf{BEV}} & \multirow{3}{*}{\textbf{Model}} & \multirow{3}{*}{\textbf{Avg.}} &
    \multicolumn{2}{c|}{\cellcolor{blue!10}\textbf{Unders.}} &
    \multicolumn{1}{c}{\cellcolor{abilityTmp!10}\textbf{Reas.}} \\
     \cmidrule(lr){4-5} \cmidrule(lr){6-6}
     \textbf{Map} & & &
     \colorbox{blue!10}{Acc}$\uparrow$ &
     \colorbox{blue!10}{RMSE}$\downarrow$ &
     \colorbox{abilityTmp!10}{Acc}$\uparrow$ \\
    \midrule
    %\multirow{3}{*}{\rotatebox[origin=c]{90}{\makecell[c]{w/o\\Map}}}
    \multirow{3}{*}{\ding{55}}
     & Qwen3-VL & 37.53 &  50.62 & 14.32 & 47.66 \\
     & Gemma-3  & 33.17 &  44.05 & 14.65 & 40.80 \\
     & Ego3D    & 27.05 &  30.67 & 15.95 & 34.53 \\
    \addlinespace[3pt]\midrule
    %\multirow{3}{*}{\rotatebox[origin=c]{90}{\makecell[c]{w/\\Map}}}
    \multirow{3}{*}{\ding{51}}
     & Qwen3-VL & 46.25  & 52.24 & 14.25 & 54.52 \\
     & Gemma-3  & 38.12 & 47.47 & 13.85 & 42.62 \\
     & Ego3D    & 31.60  & 33.82 & 15.08 & 44.46 \\
    \bottomrule
  \end{tabular}%
  }
\end{minipage}
 \vspace{-4mm}
\hfill
% ----- RIGHT: caption on top, table on bottom -----
\begin{minipage}[t]{0.32\textwidth}
  \vspace{0pt}
  % --- caption block at the top ---
  % \captionsetup{type=table}
  % \caption{Impact of Number of Views} %\textbf{ZS}: Zero-shot, \textbf{CoT}: Chain-of-Thought, \textbf{ToT}: Tree-of-Thoughts, \textbf{SC}: Self-Consistency.}
%  \label{tab:number-view}

  \vfill

  % --- table pinned to the bottom ---
  \centering
  \resizebox{\linewidth}{!}{%
  \begin{tabular}{c|l|c|c|cc|c}
    \toprule
     {\textbf{\#}} & \multirow{3}{*}{\textbf{Model}} & \multirow{3}{*}{\textbf{Avg.}} &
    \multicolumn{1}{c}{\cellcolor{red!10}\textbf{Const.}} &
    \multicolumn{2}{c|}{\cellcolor{blue!10}\textbf{Unders.}} &
    \multicolumn{1}{c}{\cellcolor{abilityTmp!10}\textbf{Reas.}} \\
    \cmidrule(lr){4-4} \cmidrule(lr){5-6} \cmidrule(lr){7-7}
     \textbf{Views} & & &
     \colorbox{red!10}{Acc}$\uparrow$ &
     \colorbox{blue!10}{Acc}$\uparrow$ &
     \colorbox{blue!10}{RMSE}$\downarrow$ &
     \colorbox{abilityTmp!10}{Acc}$\uparrow$ \\
    \midrule
    \multirow{3}{*}{0}
     & Qwen3-VL & 24.22 & 33.40 & 35.77 & 26.20 & 29.70 \\
     & Gemma-3  & 23.28 & 33.08 & 29.15 & 25.10 & 32.71 \\
     & Ego3D    & 16.24 & 26.17 & 25.79 & 25.40 & 22.16 \\
    \addlinespace[3pt]\midrule
    \multirow{3}{*}{2}
     & Qwen3-VL & 33.63 & 37.33 & 42.90 & 18.90 & 39.58 \\
     & Gemma-3  & 29.41 & 34.12 & 36.30 & 19.34 & 37.16 \\
     & Ego3D    & 21.30 & 27.87 & 28.13 & 21.05 & 28.96 \\
    \addlinespace[3pt]\midrule
    \multirow{3}{*}{4}
     & Qwen3-VL & 39.38 & 40.70 & 47.95 & 15.47 & 44.97 \\
     & Gemma-3  & 33.38 & 35.01 & 41.37 & 15.82 & 39.59 \\
     & Ego3D    & 24.85 & 29.32 & 29.79 & 17.23 & 32.67 \\
    \addlinespace[3pt]\midrule
    \multirow{3}{*}{Full}
     & Qwen3-VL & 42.24 & 42.76 & 50.62 & 14.32 & 47.66 \\
     & Gemma-3  & 35.25 & 35.55 & 44.05 & 14.65 & 40.80 \\
     & Ego3D    & 26.48 & 30.21 & 30.67 & 15.95 & 34.53 \\
    \bottomrule
  \end{tabular}%
  }
\end{minipage}

\end{table*}

\noindent \textbf{B. Impact of Geometric Grounding.} We further examine whether an explicit spatial prior can compensate for the missing geometric prior. We provide each model with a pre-constructed BEV map rendered from source annotations: HD-map-backed top-down layouts when available, and LiDAR-ground-plane projections overlaid with annotated 3D boxes otherwise. Because this input directly provides scene layout information, we exclude the \colorbox{red!10}{Const.} tasks from this ablation and treat the BEV map as a diagnostic spatial prior rather than a deployable perception module. As shown in Table~\ref{tb:all-ablation} (Middle), BEV grounding consistently improves performance across all evaluated models, increasing average accuracy by $8.27\%$ for Qwen3-VL, $4.95\%$ for Gemma-3, and $4.55\%$ for Ego3D.
%Motivated by this finding, we next test whether explicit spatial grounding can compensate for the missing geometric prior. We provide each model with a pre-constructed BEV map rendered from source annotations: HD-map-backed top-down layouts when available, and LiDAR-ground-plane projections overlaid with annotated 3D boxes otherwise. This should be interpreted as a diagnostic spatial prior rather than a deployable perception module. As shown in Table~\ref{tab:cot_method} (Right), BEV grounding yields consistent gains across all three models: Qwen3-VL improves by $4.01\%$ on average, Gemma-3 by $2.87\%$, and Ego3D by $5.12\%$, with the largest improvements appearing on reasoning tasks. The understanding error also decreases across all models, indicating that the BEV map reduces perceptual ambiguity in addition to improving downstream reasoning. These results show that, while textual prompt engineering alone is not enough, explicit spatial grounding provides the scaffold needed for stronger spatiotemporal reasoning.

\noindent
\textbf{C. Impact of the Number of Views.} To examine whether QAs in \dataset require cross-view reasoning, we evaluate each model under \textit{0-/2-/4-/full-view}. The \textit{0-view} provides only the question and answer choices, measuring language priors without visual evidence, while \textit{2-view} and \textit{4-view} provide randomly sampled partial observations. As shown in Table~\ref{tb:all-ablation} (Right), performance improves monotonically with more views, validating the cross-view constraints enforced during QA generation.

\vspace{-.5em}
\section{Conclusion \& Discussion}
\vspace{-.5em}
We introduced \textbf{\dataset}, a spatiotemporal intelligence benchmark for evaluating whether VLMs can construct coherent multi-view scene representations, reason about spatial relationships across viewpoints, infer temporal dynamics, and remain reliable across diverse driving conditions. Built from five large-scale AD datasets, \dataset contains 15.6k human-verified QA pairs spanning 20 tasks organized around \colorbox{red!10}{Const.}, \colorbox{blue!10}{Unders.}, \colorbox{abilityTmp!10}{Reas.}, and \colorbox{green!10}{Gen.}. Our evaluation of 15 representative VLMs reveals a substantial human-model gap, with \colorbox{red!10}{Const.} emerging as the clearest bottleneck. Our diagnostic studies further show that linguistic prompting does not close this gap, while explicit spatial grounding through BEV representations consistently improves performance.

\noindent
\textbf{Discussion:} These findings suggest several directions for future research. First, improving scene construction through stronger 3D-aware representation learning and multi-view fusion may help models build more reliable internal spatial representations. Second, integrating complementary sensing modalities, such as LiDAR or map priors, may improve robustness in visually degraded environments. Finally, the limited gains from linguistic-only prompting indicate a need for architectures and training paradigms that explicitly model spatial structure and temporal dynamics.

\noindent
\textbf{Impact:} By exposing the limitations of current VLMs, \dataset provides diagnostic tools that may support the development of safer and more reliable AI systems for autonomous driving. Beyond AD, the benchmark can also benefit robotics, embodied AI, and multimodal reasoning, where agents must understand spatial relationships across viewpoints and time.
% \textcolor{red}{TODO}

 %These findings suggest several important directions for future research. First, improving scene construction including 3D-aware scene representation and multi-view fusion mechanisms may help models construct more reliable internal spatial representations. Second, integrating complementary sensing modalities, such as LiDAR or map priors, may improve robustness in visually degraded environments. Finally, current prompting and reasoning techniques appear insufficient for complex spatiotemporal reasoning tasks, indicating a need for new architectures or training paradigms that explicitly model spatial structure and temporal dynamics.

% \noindent
% \textbf{Broader Impact.}
% By exposing limitations in current VLMs, the newly introduced \dataset benchmark may help guide the development of safer and more reliable AI systems for real-world driving. Beyond AD, the benchmark may also benefit research in robotics, embodied AI, and multimodal reasoning, where understanding spatial relationships across viewpoints and time is critical.

% \clearpage
{\small
\bibliographystyle{unsrt}
\bibliography{main}
}

% \newpage
% \appendix
% \newpage

% \appendix
% %  Appendix-only PDF link style (main body keeps default hyperref appearance). 
% \hypersetup{%
%   colorlinks=true,%
%   linkcolor=appendixlink,%
%   citecolor=appendixlink,%
%   urlcolor=appendixlink,%
%   filecolor=appendixlink,%
%   menucolor=appendixlink,%
%   runcolor=appendixlink,%
%   pdfborder={0 0 0}%
% }
% Drop ``Part I'' line: must run before \part{Appendix} (not only inside \input{sections/sup}).
% \renewcommand{\thepart}{}%
% \renewcommand{\partname}{}%
% \part{Appendix} %
% \parttoc %
% \clearpage

% \input{sections/sup}
% \newpage

% \input{checklist.tex}

% \clearpage
% \appendix
% \renewcommand{\thepart}{}%
% \renewcommand{\partname}{}%
% \part{Appendix} %
% \parttoc %
% \clearpage

\end{document}